\documentclass[sigconf]{acmart}
\usepackage{algorithm}  
\usepackage{algpseudocode}  
\usepackage{amsmath}  
\usepackage{graphicx}
\usepackage{subfigure}
\usepackage{diagbox}
\usepackage{multirow}
\usepackage{bm}

\newcommand{\eg}{{\it e.g.}}

\def\BibTeX{{\rm B\kern-.05em{\sc i\kern-.025em b}\kern-.08emT\kern-.1667em\lower.7ex\hbox{E}\kern-.125emX}}

\copyrightyear{2020}
\acmYear{2020}
\setcopyright{rightsretained}
\acmConference[MM '20]{Proceedings of the 28th ACM International Conference on Multimedia}{October 12--16, 2020}{Seattle, WA, USA}
\acmBooktitle{Proceedings of the 28th ACM International Conference on Multimedia (MM '20), October 12--16, 2020, Seattle, WA, USA}
\acmDOI{10.1145/3394171.3413828}
\acmISBN{978-1-4503-7988-5/20/10}

\acmSubmissionID{}

\begin{document}
	
	\title{Occluded Prohibited Items Detection: an X-ray Security Inspection Benchmark and De-occlusion Attention Module} 
	
	\author{Yanlu Wei$^{1\ast}$, \ Renshuai Tao$^{1\ast}$, \ Zhangjie Wu$^{1}$, \ Yuqing Ma$^{1}$, \ Libo Zhang$^{2}$, \ Xianglong Liu$^{1,3\dagger}$}
	\affiliation{\institution{$^{1}$ State Key Lab of Software Development Environment, \ Beihang University}}
	\affiliation{\institution{$^{2}$ Institute of Software Chinese Academy of Sciences}} 
	\affiliation{\institution{$^{3}$ Beijing Advanced Innovation Center for Big Data-Based Precision Medicine, \ Beihang University}}
	\affiliation{\institution{\{weiyanlu, rstao, zhangjiewu, mayuqing, xlliu\}@buaa.edu.cn, \ libo@iscas.ac.cn}}
	
	\thanks{$^\ast$indicates equal contribution.\\
		$^\dagger$Corresponding author.
	}

	\renewcommand{\shortauthors}{Wei and Tao, et al.}
	\renewcommand{\authors}{Yanlu Wei, Renshuai Tao, Zhangjie Wu, Yuqing Ma, Libo Zhang, Xianglong Liu}
	
	\begin{abstract}    
		Security inspection often deals with a piece of baggage or suitcase where objects are heavily overlapped with each other, resulting in an unsatisfactory performance for prohibited items detection in X-ray images. In the literature, there have been rare studies and datasets touching this important topic. In this work, we contribute the first high-quality object detection dataset for security inspection, named Occluded Prohibited Items X-ray (OPIXray) image benchmark. OPIXray focused on the widely-occurred prohibited item "cutter", annotated manually by professional inspectors from the international airport. The test set is further divided into three occlusion levels to better understand the performance of detectors. Furthermore, to deal with the occlusion in X-ray images detection, we propose the De-occlusion Attention Module (DOAM), a plug-and-play module that can be easily inserted into and thus promote most popular detectors. Despite the heavy occlusion in X-ray imaging, shape appearance of objects can be preserved well, and meanwhile different materials visually appear with different colors and textures. Motivated by these observations, our DOAM simultaneously leverages the different appearance information of the prohibited item to generate the attention map, which helps refine feature maps for the general detectors. We comprehensively evaluate our module on the OPIXray dataset, and demonstrate that our module can consistently improve the performance of the state-of-the-art detection methods such as SSD, FCOS, etc, and significantly outperforms several widely-used attention mechanisms. In particular, the advantages of DOAM are more significant in the scenarios with higher levels of occlusion, which demonstrates its potential application in real-world inspections. The OPIXray benchmark and our model are released at {\url{https://github.com/OPIXray-author/OPIXray}}.  
	\end{abstract}


	\fancyhead{}
	\maketitle 
	
	\begin{figure}[h]
		\centering
		\includegraphics[width=\linewidth]{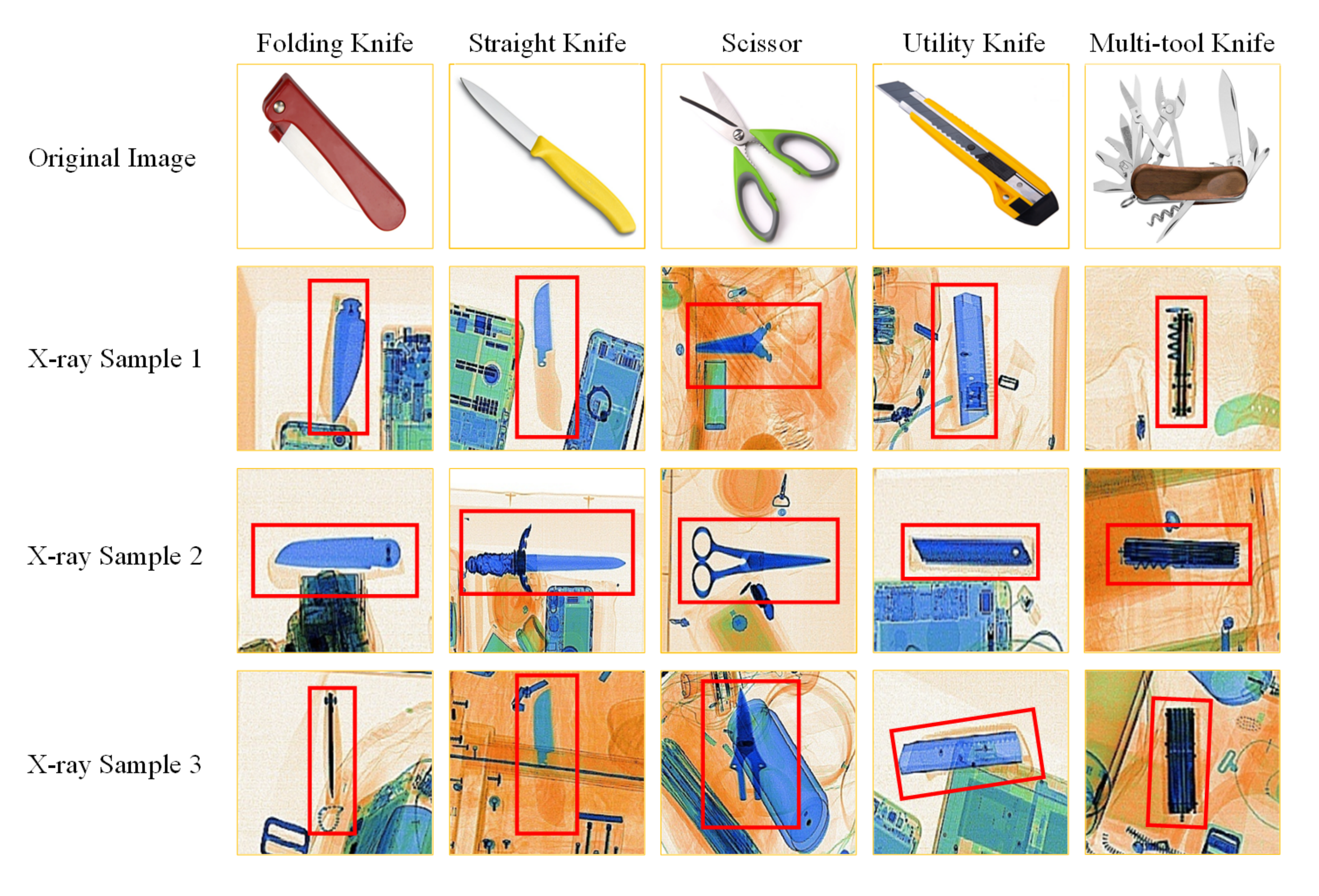}
		\caption{Samples of the five categories of cutters and corresponding X-ray images.}
		\Description{knife-classes}\label{Samples_five}
	\end{figure}
	
	\section{Introduction}
	With the increasing crowd density in public transportation hubs, security inspection has become more and more important in protecting public safety. Security inspection usually adopts X-ray scanners to find whether there is any prohibited item in passenger luggage. In this scenario, objects in the luggage are randomly stacked and heavily overlapped with each other, leading to heavy object occlusion. As a result, after a long time localizing prohibited items in large amounts of complex X-ray images without distraction, security inspectors struggle to accurately detect all the prohibited items, which may cause severe danger to the public. And changing shifts frequently will cost a large number of human resources, which is not advisable. 
	
	Therefore, a rapid, accurate and automatic approach to assist inspectors to detect prohibited items in X-ray scanned images is desired eagerly. As the technology of deep learning \cite{lecun2015deep} especially the convolutional neural network develops \cite{wang2016action,cheng2019improving}, the recognition of occluded prohibited items from X-ray pictures can be regarded as an object detection problem of computer vision, which has been widely studied in the literature.
	
	There are several works trying to solve occlusion problems in different scenarios, such as person re-identification \cite{zhang2018occlusion,zhou2018bi,wang2018repulsion}, face recognition \cite{ge2017detecting,yang2016nuclear,song2019occlusion}. The object occlusion in Person Re-identification or Face Recognition belongs to intra-class occlusion, and every occluded object has a corresponding annotation. Therefore, a loss function can be designed by annotation information to decrease the impact of occlusion. However, object occlusion in X-ray images for security inspection often exists between prohibited items and safety items, which belongs to inter-class occlusion. For the prohibited items detection task, what we obtained are the annotations of prohibited items, so methods of these senses can not be used for comparison. To the best of our knowledge, up to now, no dataset targeting occluded prohibited items detection in X-ray images has been proposed by researchers even there are also two released X-ray benchmark, namely GDXray \cite{mery2015gdxray} and  SIXray \cite{miao2019sixray}. However, GDXray \cite{mery2015gdxray} contains images which are grayscale, while another dataset SIXray \cite{miao2019sixray} only contains less than 1\% images having annotated prohibited items. And both GDXray \cite{mery2015gdxray} and SIXray \cite{miao2019sixray} are used for classification task. As a result, both of the two datasets are inconsistent with our task that detecting occluded prohibited items.
	
	To torch this important topic, we contribute the first high-quality object detection dataset for security inspection, named Occluded Prohibited Items X-ray (OPIXray) image benchmark. Considering that cutter is the most common tool passengers carry, we choose it as the prohibited item to detect. OPIXray contains 8885 X-ray images of 5 categories of cutters (illustrated in Fig. \ref{Samples_five}). Each sample has at least one prohibited item, while some samples have more. All samples are annotated manually by the professional inspectors from the international airport and the standard of annotating is based on the standard of training security inspectors. Our dataset brings meaningful challenges to this topic in two main folds. First, OPIXray mimics a similar testing environment to the real-world scenario, where items randomly overlapped with each other, leading to object occlusion challenge. Second, cutters of different categories usually share the similar shape appearance, \eg, folding knives and multi-functional knives, bringing difficulties to discriminate.
	
	Furthermore, to deal with the occlusion in X-ray images, we propose the De-occlusion Attention Module(DOAM), a plug-and-play module that can be easily inserted into most popular detectors. As we have observed, X-ray imaging preserves the shape appearance in the heavy occluded part and assigns various colors to different materials in the visual part. Inspired by the fact we observed, our module simultaneously lays particular emphasis on edge information and material information of the prohibited item by utilizing two sub-modules, namely, Edge Guidance (EG) and Material Awareness (MA). Then, our module leverages the two information above to generate an attention distribution map as a high-quality mask for each input sample to generate high-quality feature maps, serving identifiable information for the general detectors.
	
	The main contributions of this work are as follows:
	\begin{itemize}
		\item We provide the first benchmark for occluded prohibited items detection in X-ray images for security inspection. The OPIXray dataset we contributed is high-quality because all prohibited items are manually annotated by professional security inspectors from the international airport.
		\item We present the De-occlusion Attention Module (DOAM), simultaneously laying particular emphasis on edge information and material information of the prohibited item, inspired by the X-ray imaging principle.
		\item DOAM can be easily inserted as a plug-and-play module into various detectors, including SSD \cite{liu2016ssd}, YOLOv3 \cite{redmon2018yolov3} and FCOS \cite{tian2019fcos}, etc., which means our module can be widely applied.
		\item We evaluate our method on the OPIXray dataset and compare it to various baselines, including popular detection approaches and widely-used attention mechanisms. These results show that DOAM can not only consistently improve the performance of the state-of-the-art detection methods but also significantly outperform several widely-used attention mechanisms.
		
	\end{itemize}
	
	\section{Related Work}
	\subsection{X-ray Images and Benchmarks} 
	X-ray offers powerful ability in many tasks such as medical imaging analysis \cite{guo2019improved,chaudhary2019diagnosis,lu2019towards} and security inspection \cite{miao2019sixray,huang2019modeling}. However, the visibility of the object information contained in X-ray images suffers a lot because of object occlusion.
	
	Several studies in the literature have attempted to address this challenging problem. Unfortunately, due to the particularity of security inspection, very few X-ray datasets have been published. A released benchmark, GDXray\cite{mery2015gdxray} contains 19407 images, partial of which contains three categories of prohibited items including gun, shuriken and razor blade. However, GDXray only contains gray-scale images in a very simple background, which is 
	far away from real-world scenario. Recently, SIXray\cite{miao2019sixray} is a large-scale X-ray dataset which is about 100 times larger than the GDXray dataset\cite{mery2015gdxray}. SIXray consists of 1059231 X-ray images, but the positive samples are less than 1\% to mimic a similar testing environment to the real-world scenario where inspectors often aim at recognizing prohibited items appearing in a very low frequency. Different from ours, SIXray is a dataset for the task of classification, focusing on the problem of data imbalance. 
	\subsection{Attention Mechanism} 
	
	Attention can be interpreted as a means of biasing the allocation of available computational resources towards the most informative components of a signal, which has been widely studied in many tasks, like image retrieval \cite{yao2019attention,sun2019attention}, visual question answering \cite{peng2019cra,zha2019spatiotemporal}. It captures long-range contextual information and has been widely applied in various tasks such as machine translation \cite{vaswani2017attention}, image captioning \cite{chen2017sca}, scene segmentation \cite{fu2019dual} and object recognition \cite{tang2015rgb}. The work \cite{wang2018non} is related to self-attention module, mainly exploring the effectiveness of non-local operation in space-time dimensions for videos and images. \cite{fu2019dual} proposed a dual attention network (DANet) for scene segmentation by capturing contextual dependence based on the self-attention mechanism. Squeeze-and-Excitation Networks (SENet) \cite{hu2018squeeze} terms the Squeeze-and-Excitation block (SE), that adaptively re-calibrates channel-wise feature responses by explicitly modeling inter-dependencies between channels.
	
	\begin{table}[h] 
		\setlength{\tabcolsep}{3pt}
		\caption{The category distribution of the OPIXray dataset. Due to that some images contain more than one prohibited item, the sum of all items in the different categories is greater than the total number of images.}
		\label{data_table}
		\begin{tabular}{c|ccccc|c}
			\hline
			\multirow{2}{*}{OPIXray} & \multicolumn{5}{c|}{Categories}                     & \multirow{2}{*}{Total} \\ \cline{2-6}
			& Folding & Straight & Scissor & Utility & Multi-tool &                        \\ \hline
			Training                     & 1589    & 809      & 1494    & 1635    & 1612       & 7109                   \\
			Testing                      & 404     & 235      & 369     & 343     & 430        & 1776                   \\ \hline
			Total                            & 1993    & 1044     & 1863    & 1978    & 2042       & 8885                   \\ \hline
		\end{tabular}
	\end{table}
	
	\begin{table}[t]
		\setlength{\tabcolsep}{2pt}
		\caption{The category distribution of different occlusion levels in the testing set.}
		\label{data_cate_table}
		\begin{tabular}{c|ccccc|c}
			\hline
			\multirow{2}{*}{Testing set} & \multicolumn{5}{c|}{Categories}                     & \multirow{2}{*}{Total} \\ \cline{2-6}
			& Folding & Straight & Scissor & Unility & Multi-tool &                        \\ \hline
			OL1                          & 206     & 88       & 160     & 214     & 255        & 922                    \\
			OL2                          & 148     & 84       & 126     & 88      & 105        & 548                    \\
			OL3                          & 50      & 63       & 83      & 41      & 70         & 306                    \\ \hline
			Total                        & 404     & 235      & 369     & 343     & 430        & 1776                   \\ \hline
		\end{tabular}
	\end{table}
	
	\section{The OPIXray Dataset}
	The performance of deep learning models largely depends on the quality of the dataset. Only with high quality dataset can the detection ability of a model be evaluated reasonably. Thus, a professional dataset with high-quality annotations is necessary for training models and performing evaluations. In this work, we build the first dataset specially designed for occluded prohibited items detection in security inspection.
	\subsection{Data properties}
	\textbf{Data Acquisition:} The backgrounds of all samples are scanned by the security inspection machine and the prohibited items are synthesized into these backgrounds by the professional software. In the international airport, these synthesized images are used to train security inspectors to recognize prohibited items, which is exactly what we want to be executed automatically. And each prohibited item is annotated manually by professional inspectors from the international airport, which localized by a box-level annotation with a bounding box. These X-ray images still retain the specific property that X-ray imaging preserves the shape appearance in the heavy occluded part and assigns various colors to different materials, mainly reflected in the visual part. 
	
	\textbf{Data Structure:} The OPIXray dataset contains a total of 8885 X-ray images of 5 categories of cutters, namely, Folding Knife, Straight Knife, Scissor, Utility Knife, Multi-tool Knife. A statistics of category distribution is shown in Tab. \ref{data_table}. All images are stored in JPG format with the resolution of 1225*954. The dataset is partitioned into a training set and a testing set, with the former containing 80\% of the images (7109) and the latter containing 20\% (1776), where the ratio is about 4 : 1. The statistics of category distribution of training set and testing set are also shown in Tab. \ref{data_table}. Note that there are 35 images of the dataset, each of which contains more than one prohibited item, by 30 in the training set and 5 in the test set.
	
\begin{figure}[b]
		\centering
		\includegraphics[width=\linewidth]{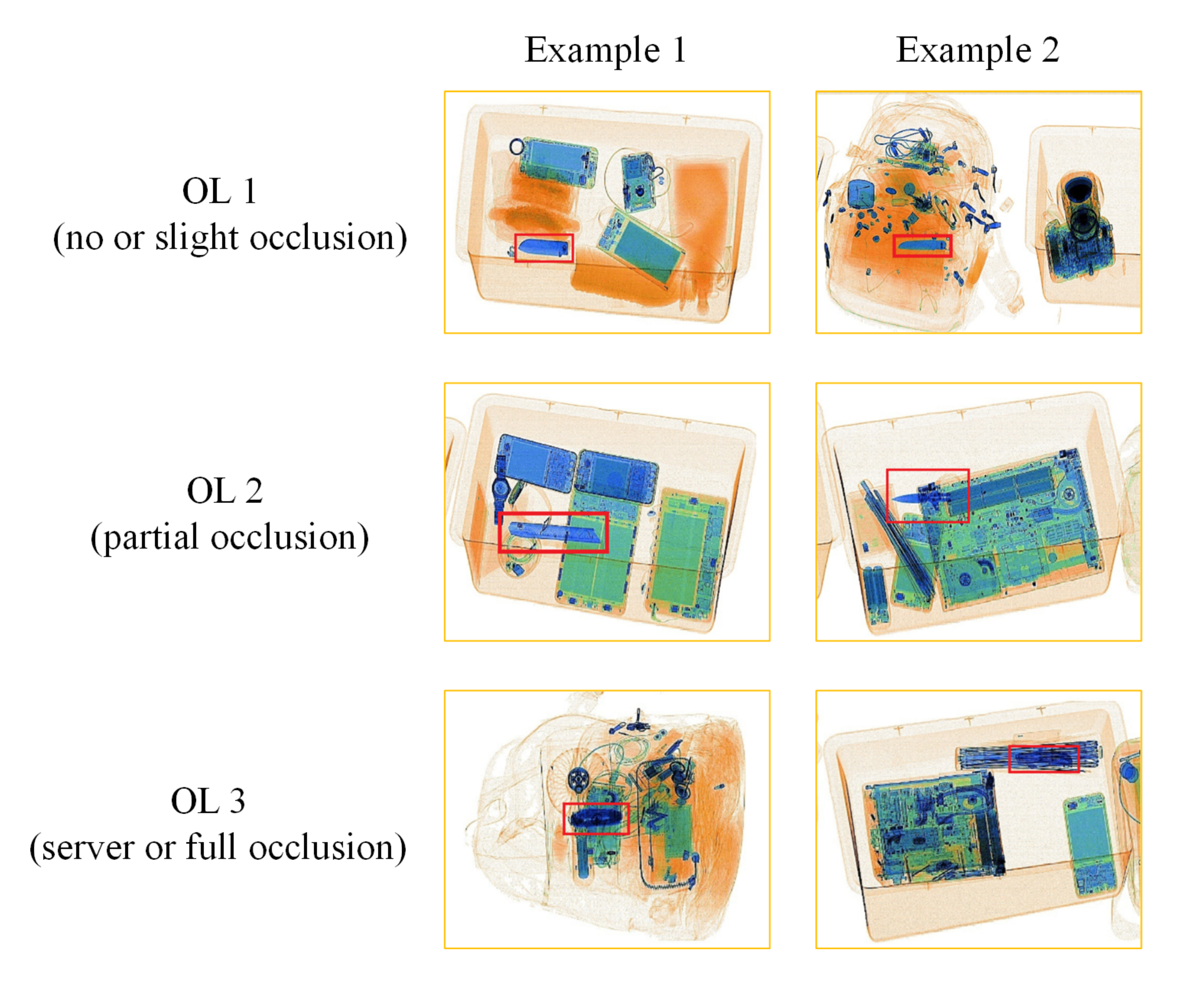}
		\caption{Samples of different occlusion levels.}\label{Test_Occ}
		\Description{}
	\end{figure}

	\textbf{Data Occlusion Levels:} In order to study the impact brought by object occlusion levels, we divide the testing set into three subsets and name them Occlusion Level 1 (OL1), Occlusion Level 2 (OL2) and Occlusion Level 3 (OL3), where the number indicates occlusion level of prohibited items in images. As illustrated in Fig. \ref{Test_Occ}, there is no or slight occlusion on prohibited items in OL1 and partial occlusion in OL2. To maximally evaluate the ability of models to deal with object occlusion, we construct OL3 by choosing images where severe or full occlusion exists in. The category distribution of the three subsets with different occlusion levels are shown in Tab. \ref{data_cate_table}. 
	
	\subsection{Dataset Analysis}
	\textbf{Data Authenticity:} The OPIXray dataset mostly mimics a similar environment to the real-world scenario. \textbf{First}, the occlusion of prohibited items is inspired that items within personal luggage are usually stacked randomly and overlapped with each other, which we describe in detail in this work. \textbf{Second}, the statics of category distribution is inconsistent obviously. The number of folding knife and multi-tool knife are higher than straight knife because the former two categories are more common for passengers to bring. And the number of OL3 is significantly less than OL1 because cutters are usually small and move freely in luggage, as a result, cutters are seldom fully occluded in the real scenario.
	
\begin{figure}[htbp]
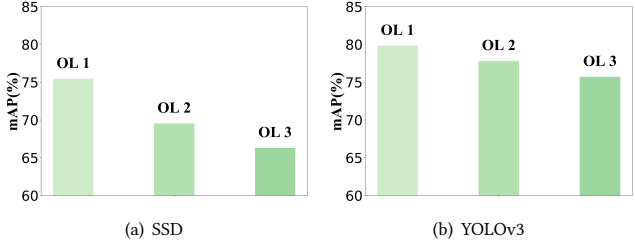

		\centering
		\subfigure[SSD]{
			\begin{minipage}[t]{0.5\linewidth}
				\centering
				\includegraphics[width=1.6in]{ssd.pdf}
		\end{minipage}}%
		\subfigure[YOLOv3]{
			\begin{minipage}[t]{0.5\linewidth}
				\centering
				\includegraphics[width=1.6in]{yolo3.pdf}
		\end{minipage}}%
		\centering
		\caption{The performance of SSD and YOLOv3 under three different object occlusion levels.} \label{data_analyses} 
	\end{figure}

	\textbf{Data Application:} OPIXray dataset has two major application scenarios. \textbf{First}, the dataset can evaluate the ability of a model to detect prohibited items in X-ray images. A better model can achieve better performance no matter which occluded levels. As we can see from Fig. \ref{data_analyses}, there is a significant decline in the performance of famous detection approaches \eg, SSD \cite{liu2016ssd} and YOLOv3 \cite{redmon2018yolov3}, with the occlusion level increasing. \textbf{Second}, the dataset can evaluate the ability of a model of solving object occlusion problem, by comparing the improvement than other methods in different occlusion level settings. The improvement amount of an approach increases with the occlusion level increases, which illustrates the effectiveness of this approach to the object occlusion problems.
	
	\section{De-occlusion Attention Module}
	We propose the De-occlusion Attention Module(DOAM), simultaneously laying particular emphasis on edge information and material information of the prohibited item by utilizing two sub-modules, namely, Edge Guidance (EG) and Material Awareness (MA). Then, our module leverages the two information above to generate an attention distribution map as a high-quality mask for each input sample to generate high-quality feature maps, serving identifiable information for general detectors. Without losing of generality, we apply the DOAM to the widely-used SSD \cite{liu2016ssd} and demonstrate our design from the following aspects: 1) how DOAM work briefly (4.1); 2) how to impel the edge information to guide the model (4.2); 3) how to aggregate the region information to express the material information (4.3); 4) how to leverage the two source of information and generate the attention map (4.4); 5) how to compare our module with the base detector and other counterparts (4.5).
	
	\subsection{Network Architecture}
	
\begin{figure*}[h]
		\centering
		\includegraphics[width=\linewidth]{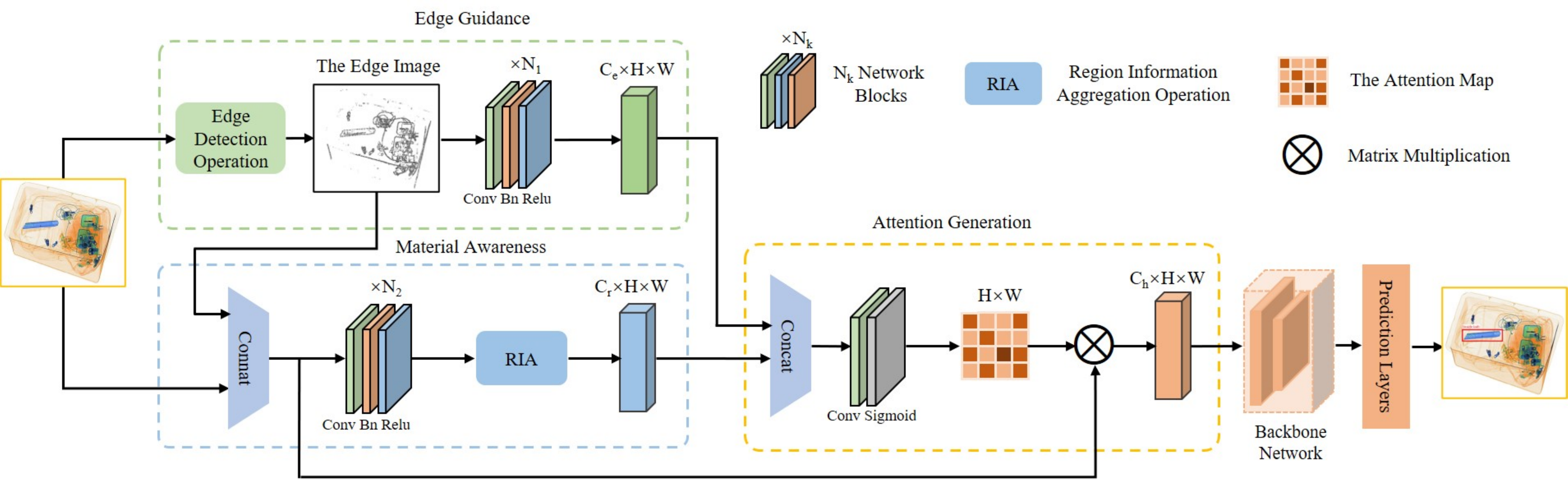}
		\caption{DOAM integrated with a general backbone network architecture. As illustrated, two feature maps are generated by Edge Guidance (EG) and Material Awareness (MA) and fused to generate the attention map in Attention Generation. Further, the attention map is applied to the input image to generate refined feature maps we desire. Finally, the refined feature map can be utilized by the SSD network.}
		\Description{Network architecture of the SSD with our  DOAM.}\label{framework}
	\end{figure*}

	Fig. \ref{framework} illustrates the architecture of the SSD detector with the proposed DOAM.
	On the top of the SSD, DOAM leverages edge and material information generated by two parallel branches, namely EG and MA, to generate an attention distribution map, providing enhanced features for further accurate detection.

	Specifically, suppose there are $n$ training images in the dataset $\LARGE{X} = \left\{\mathbf{x}_1,\cdots,\mathbf{x}_n\right\}$. Each input image $\mathbf{x}\in\LARGE{X}$ is fed into EM and MA to obtain $\mathbf{F}_{E}$ and $\mathbf{F}_{M}$, laying particular emphasis on edge information of the occluded part and material information of the visual part, respectively. In EG, we first extract the edge map through an edge detection operation and then generate the edge guidance information $\mathbf{F}_{E}$ which emphasizes the complete edge information of the prohibited item, especially in occluded part.
	In MA, we take the concatenation of $\mathbf{x}$ and edge guidance $\mathbf{F}_{E}$ as input and denote it as $\mathbf{P}$, and extract a temporary feature map $\mathbf{F}_{tmp1}$ (Note that $\mathbf{F}_{tmp1}$, $\mathbf{F}_{tmp2}$ and $\mathbf{F}_{tmp3}$ are
	intermediate states of the refined feature $\mathbf{F}_{M}$ during the refining process). To further aggregate the region information to emphasis the material characteristics of the input image, we design a Region Information Aggregation (RIA) operation where different pooling kernels are utilized to aggregate multi-scale region-wise features which will be selected by a gated CNN to further adaptively generate the material awareness information $\mathbf{F}_{M}$. $\mathbf{F}_{M}$ remains and emphasizes information of identifiable properties of the visible part. Finally, We fuse the edge guidance information $\mathbf{F}_{E}$ and material awareness information $\mathbf{F}_{R}$ to obtain the attention distribution map $\mathbf{S}$. With the help of $\mathbf{S}$, we can obtain the enhanced features $\mathbf{F}$ of the input image for further accurate detection.
	The entire process of DOAM is illustrated in detail in Algorithm \ref{Alg}.
	
	\begin{algorithm}[t]
		\caption{The Operation Process of DOAM.} \label{Alg}
		\begin{algorithmic}[1]
			\State \textbf{Input}: an X-ray image $\mathbf{x}\in\mathbb{R}^{C\times{H}\times{W}}$;
			\State Generate the horizontal edge image $\mathbf{E}^{h}$ and the vertical edge image $\mathbf{E}^{v}$ by the $Sobel$ operator.
			\State Generate the edge image $\mathbf{E}$ by synthesizing $\mathbf{E}^{h}$ and $\mathbf{E}^{v}$.
			\For {$N_1$ steps}
			\State Refine the feature map $\mathbf{F}_{E}$ by operating $\mathbf{E}$ through $f_{\mathbf{e}}(\cdot)$.
			\EndFor
			\State Generate the concatenated image $\mathbf{P}$ by concatenating $\mathbf{x}$ and $\mathbf{E}$.
			\For {$N_2$ steps}
			\State Refine the feature map $\mathbf{F}_{tmp1}$ by operating $\mathbf{P}$ through $f_{\mathbf{r}}(\cdot)$.
			\EndFor
			\For {$k\in\left\{\mathbf{k}_1,\ldots,\mathbf{k}_n\right\}$}
			\State Generate refined feature map $\mathbf{F}_{tmp2}^k$ by operating $\mathbf{F}_{tmp1}$ through Eq. (\ref{B}).
			\State Generate refined feature map $\mathbf{F}_{tmp3}^k$ by concatenating $\mathbf{F}_{tmp1}$ and $\mathbf{F}_{tmp2}^k$.
			\State Update the feature map set $\LARGE{S}$ = $\LARGE{S} \cup \mathbf{F}_{tmp3}^k$.
			\EndFor
			\State Choose the appropriate feature map $\mathbf{F}_{M}$ from $\LARGE{S}$ by drawing the gated convolutional network.
			\State Generate the fused feature map $\mathbf{F}_{fus}$ by operating $\mathbf{F}_{E}$ and $\mathbf{F}_{M}$.
			\State Generate the attention map $\mathbf{S}=\sigma(\mathbf{F}_{fus})$.
			\State Generate the final feature map $\mathbf{F}$ by performing a matrix multiplication between $\mathbf{S}$ and $\mathbf{P}$.
			\State \textbf{Output}: the refined feature map $\mathbf{F}\in\mathbb{R}^{C_h\times{H}\times{W}}$.
		\end{algorithmic}
	\end{algorithm}
	
	\subsection{Edge Guidance (EG)}
	
	For each input image $\mathbf{x}\in\LARGE{X}$, we utilize the convolutional neural network with the horizontal and vertical kernel denoted as $s_h$, $s_v$ of the $Sobel$ operator, to respectively compute the edge images $\mathbf{E}^{h}$ and $\mathbf{E}^{v}$ in horizontal and vertical directions. We further generate the edge image $\mathbf{E}$ of the input image $\mathbf{x}$ by synthesizing the above two results $\mathbf{E}^{h}$ and $\mathbf{E}^{v}$. To lead EG to only magnify edge information of the prohibited items, we use $\mathbf{N}_1$ network blocks (Here, we define $\mathbf{N}_1$ as the Module Operation Intensity of EG, which represents that the performance of the module changes with the value of $\mathbf{N}_1$.), in which each block consists of a convolutional layer with a $3\times3$ kernel size, a batch normalization layer, and a relu layer, to extract the feature map $\mathbf{F}_{E}$. The operations can be formulated as follows:    
	\begin{equation}
	f_{\mathbf{e}}(\mathbf{a})=relu\left(\mathbf{W}_{e}\cdot\mathbf{a}+\mathbf{b}_{e}\right).
	\end{equation}
	\begin{equation}\label{F:EG} 
	\mathbf{F}_{E}=\left\{f_{\mathbf{e}}(\mathbf{E})\right\}_{N_1}.
	\end{equation}
	where $\left\{\cdot\right\}_{N_1}$ means that the operation is repeated $N_1$ times, $\mathbf{W}_{e}$, $\mathbf{b}_{e}$ are parameters of the convolutional layer. After extracting the feature map $\mathbf{F}_{E}$ as shown in Eq. \ref{F:EG}, we adaptively attend to the edge guidance information of the prohibited item within the feature map $\mathbf{F}_{E}$ by optimization.

	\subsection{Material Awareness (MA)}
	
	Material information is mainly reflected in color and texture. For color information, each position of the image has the ability to represent. However, when it comes to the texture information, each position needs to combine its surroundings to represent. Inspired by the fact that the aggregation of regional information can represent both color and texture, we define that the region information after aggregated is the representation of the material information. In order to construct relations between each position of the concatenated image (the input image $\mathbf{x}$  and its edge image $\mathbf{E}$ in EG) and a certain region around the point, we utilize $\mathbf{N}_2$ network blocks (As we state $\mathbf{N}_1$ in EG, $\mathbf{N}_2$ is the Module Operation Intensity of MA.), in which each block consists a convolutional layer with the kernel size is $3\times3$, a batch normalization layer, a relu layer, to extract a temporary feature map $\mathbf{F}_{tmp1}$ from the concatenated image as follows:
	\begin{equation}
	f_{\mathbf{r}}(\mathbf{a})=relu\left(\mathbf{W}_{r}\cdot\mathbf{a}+\mathbf{b}_{r}\right).
	\end{equation}
	\begin{equation}\label{A} 
	\mathbf{F}_{tmp1}=\left\{f_{\mathbf{r}}(\mathbf{x}||\mathbf{E})\right\}_{N_2}.
	\end{equation}
	where $||$ represents concatenating operation. We further generate the refined feature map $\mathbf{F}_{M}$ of MA by refining $\mathbf{F}_{tmp1}$ through the Region Information Aggregation (RIA) operation, central of MA. 
	
\begin{figure}[t]
		\centering
		\includegraphics[width=\linewidth]{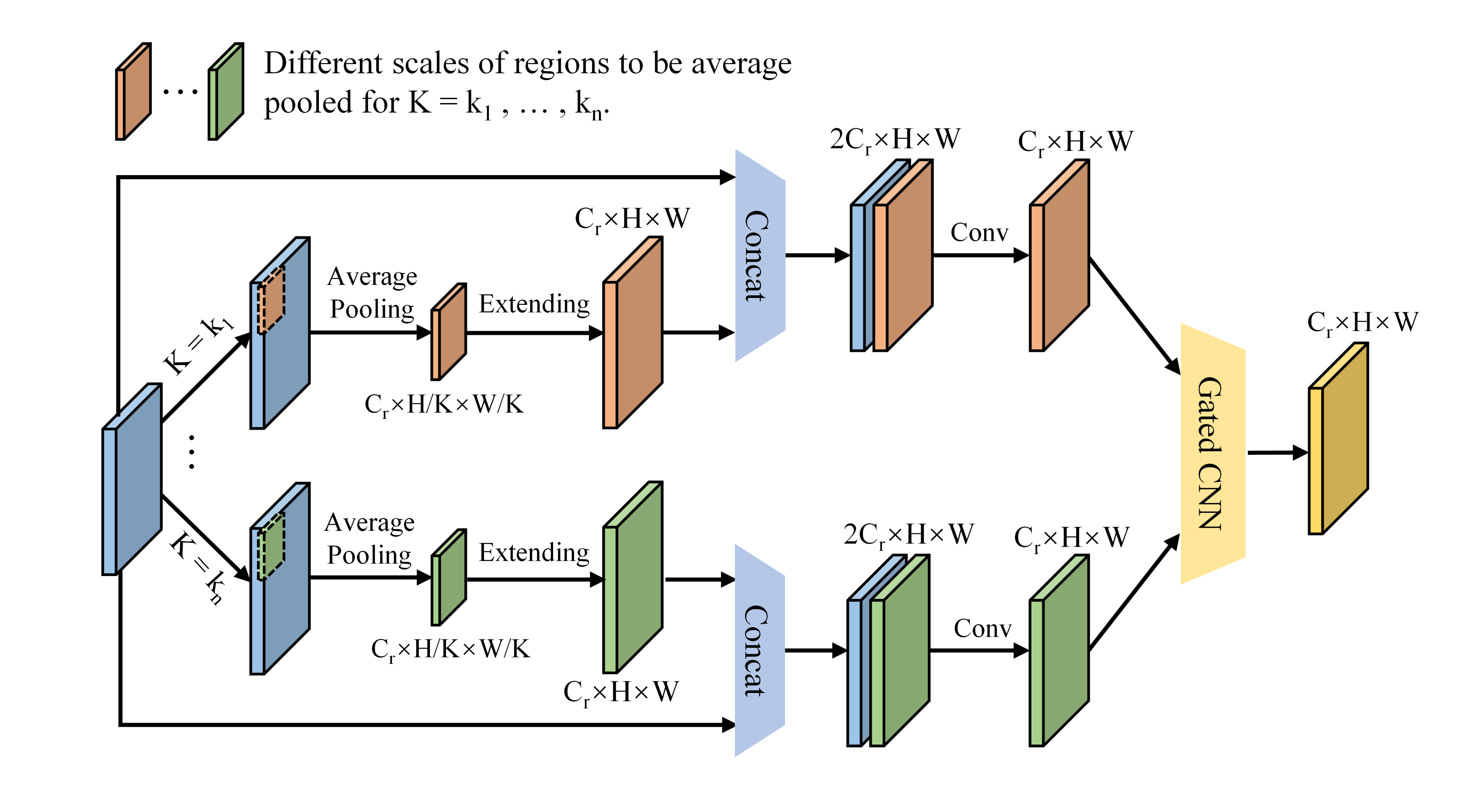}
		\caption{The Operation Process of RIA.}
		\Description{The Operation Process of RIA.}\label{RIA}
	\end{figure}

	Fig. \ref{RIA} illustrates the detailed process of RIA operation. For the input feature map $\mathbf{F}_{tmp1}$ and a parameter $k$, RIA operation aggregates the information of a certain size of $k\times k$ of region around it by average pooling and extending to generate another temperate feature map $\mathbf{F}_{tmp2}^k$.
	The average pooling and extending operations can be formulated together as follows:
	\begin{equation}\label{B}
	{\mathbf{F}_{tmp2}}^k_{ij}=\frac{\sum _{m=i-(i\bmod{k})}^{i-(i\bmod{k})+k}\sum _{n=j-(j\bmod{k})}^{j-(j\bmod{k})+k}{\mathbf{F}_{tmp1}}_{mn} }{k^2}.
	\end{equation}
	where ${\mathbf{F}_{tmp2}}^k_{ij}$ represents the feature of the $i$-th row and $j$-th column of feature map $\mathbf{F}_{tmp2}$ when the kernel size for the average pooling layer is $k$.
	
	We further concatenate the two feature maps ($\mathbf{F}_{tmp1}$ and $\mathbf{F}_{tmp2}$) in the dimension of channel to generate a new feature map $\mathbf{F}_{tmp3}$, where the dimensions are $2C_r\times{H}\times{W}$.
	Then every point of the new feature map has the ability to perceive the region of size $k\times k$ around it, which means that the relations have been constructed. Due to different sizes of region information to aggregate (different values of $k$), the module generates a feature map set $\LARGE{S}$=$\left\{\mathbf{F}_{tmp3}^{k_1},\cdots,\mathbf{F}_{tmp3}^{k_n}\right\}$.
	
	In order to enable RIA operation to perform well on various scales of prohibited items, it is necessary to design a mechanism to adaptively choose an optimal value for $k$. We exploit the gated convolutional neural network \cite{yu2019free} $\mathbb{G}$ with $3\times3$ kernels into RIA, to select the proper feature map $\mathbf{F}_{M}$ from the feature map set $\LARGE{S}$ as output. The operations are formulated as follows:
	\begin{equation}\label{F:MP}
	\mathbf{F}_M = \mathbb{G}(\LARGE{S}).
	\end{equation}
	where $\LARGE{S}$=$\left\{\mathbf{F}_{tmp3}^{k_1},\cdots,\mathbf{F}_{tmp3}^{k_n}\right\}$. 
	
	\subsection{Attention Generation}
	As is illustrated in Algorithm \ref{Alg}, for the result feature maps $\mathbf{F}_{E}$ and $\mathbf{F}_{M}$ outputted by the EG and the MA respectively, where $\mathbf{F}_{E}\in \mathbb{R}^{C_e\times{H}\times{W}}$, $\mathbf{F}_{M}\in \mathbb{R}^{C_r\times{H}\times{W}}$, we concatenate them for information fusion. And further we feed the concatenated feature into a convolutional layer, where the kernel size is 1$\times$1, to generate the feature map $\mathbf{F}_{fus}\in \mathbb{R}^{(C_e+C_r)\times{H}\times{W}}$, which have confused the edge information and the reginal information, both strengthened. The operation can be formulated as follows: 
	\begin{equation}\label{F:mix}
	\mathbf{F}_{fus}=\mathbf{W}_{m}\left(\mathbf{F}_{E}||\mathbf{F}_{M}\right)+\mathbf{b}_{m}.
	\end{equation}
	where $||$ represents the operation of concatenating, and $\mathbf{W}_{m}$, $\mathbf{b}_{m}$ are parameters of the convolutional layer. Then we utilize the feature map $\mathbf{F}_{fus}$ as the input of a sigmoid function to generate the attention map $\mathbf{S}$ as follows:
	\begin{equation}\label{S}
	\mathbf{S}=\sigma(\mathbf{F}_{fus}) = \frac{1}{1+e^{-\mathbf{F}_{fus}}}.
	\end{equation}
	where $\mathbf{S}\in\mathbb{R}^{{H}\times{W}}$. Finally, we calculate the inner product of the attention map $\mathbf{S}$ and the concatenated image $\mathbf{P}$ as follows:
	\begin{equation}\label{M}
	\mathbf{F}_{j}=\sum_{i=1}^{H\times{W}}\mathbf{S}_{ji}\mathbf{P}_i. 
	\end{equation}
	where $\mathbf{F}\in\mathbb{R}^{{C}_{h}\times{H}\times{W}}$, and it is the final refined feature map we desire to serve to detectors, of which the information highly contributes to the detection of the prohibited item are emphasised.
	
	\subsection{Module Complexity Analysis}
	
	In this section, we analyze the model complexity with or without DOAM in SSD \cite{liu2016ssd} and compare the complexity including the total number of parameters, model size and computation cost, with other attention mechanisms.
	
	Table \ref{param-table} reports that DOAM only brings a slight increase in computational cost (7.14\% in GFLOPs), compared to the SSD \cite{liu2016ssd} without any attention mechanisms. Additionally, we compare the complexity between DOAM and three variants of attention mechanisms, including SE \cite{hu2018squeeze}, Non-local \cite{wang2018non} and DA \cite{fu2019dual}. The three attention mechanisms focus on channel information, spatial information and combination of the two kinds of information, respectively. As we can see from table \ref{param-table}, compared to the single SSD \cite{liu2016ssd}, \textbf{First}, for total number of parameters, SE \cite{hu2018squeeze}, Non-local \cite{wang2018non} and DA \cite{fu2019dual} respectively bring 32.23\%, 27.69\% and 88.43\% increases, while the increase our module brings is almost negligible. \textbf{Second}, for model size, SE \cite{hu2018squeeze}, Non-local \cite{wang2018non} and DA \cite{fu2019dual} respectively bring 31.86\%, 27.32\% and 88.01\% increases, while the increase our module brings is almost negligible. However, for computational cost, SE \cite{hu2018squeeze}, Non-local \cite{wang2018non} and DA \cite{fu2019dual} respectively bring 3.15\%, 6.54\% and 23.07\% increases, while our module brings 7.14\%. 
	
	In conclusion, for total number of parameters and model size, DOAM is much more computation efficient than the three famous attention mechanisms. For computational cost, DOAM is slightly more computationally expensive. We conjecture that it is mainly because that different values of parameter $k$ cause repetitive computation in RIA.
	
	\begin{table}[h]
		\setlength{\tabcolsep}{8pt}
		\caption{Complexity comparison of different models. PARAMs, SIZE and GFLOPs represent the total number of parameters, the Model Size and the Giga Floating Point operations, respectively.}
		\label{param-result}
		\begin{tabular}{lccc}
			\toprule
			Method & PARAMs & SIZE(MB) & GFLOPs\\
			\hline
			SSD \cite{liu2016ssd} & $24.2\times10^6$ & 92.6 & 30.6522\\
			\midrule
			SSD+SE \cite{hu2018squeeze}  &   $32.0\times10^6$  &  122.1 & \textbf{31.6169}\\
			SSD+Non-local \cite{wang2018non}  & $30.9\times10^6$ & 117.9 &  32.6577\\
			SSD+DA \cite{fu2019dual} & $45.6\times10^6$ & 174.1 & 37.7231\\      
			\textbf{SSD+DOAM(ours)} & $\bm{24.3\times10^6}$ & \textbf{92.7} & 32.8435\\
			\bottomrule
		\end{tabular}
		\label{param-table}
	\end{table}
	
	\begin{table}[h]
		\setlength{\tabcolsep}{3pt}
		\caption{Performance comparison between DOAM and other different attention mechanisms on object categories. FO, ST, SC, UT and MU represent Folding Knife, Straight Knife, Scissor, Utility Knife and Multi-tool Knife, respectively.}   
		\label{Classification-result}
		\begin{tabular}{lcccccc}
			\toprule
			\multicolumn{1}{l}{\multirow{2}{*}{Method}} & \multicolumn{1}{c}{\multirow{2}{*}{mAP}} & \multicolumn{5}{c}{Categories}                                                        \\ \cline{3-7} 
			\multicolumn{1}{l}{} 
			& \multicolumn{1}{c}{}
			& \multicolumn{1}{c}{FO} & \multicolumn{1}{c}{ST} & \multicolumn{1}{c}{SC} & \multicolumn{1}{c}{UT} & \multicolumn{1}{c}{MU} \\ \hline
			SSD \cite{liu2016ssd} & 70.89  & 76.91  & 35.02  & 93.41    & 65.87& 83.27\\
			\hline
			SSD+SE \cite{hu2018squeeze}  & 71.85  & 77.17 & 38.29  & 92.03  & 66.10   & \textbf{85.67} \\
			SSD+Non-local \cite{wang2018non}  & 71.41 & 77.55 & 36.38    & \textbf{95.26}   & 64.86   & 82.98 \\
			SSD+DA \cite{fu2019dual} & 71.96   & 79.68 & 37.69    & 93.38  & 64.14  & 84.90\\      
			\textbf{SSD+DOAM(ours)} &\textbf{74.01} & \textbf{81.37} & \textbf{41.50} & 95.12 & \textbf{68.21} & 83.83\\
			\bottomrule
		\end{tabular}\label{table_attention_catagories}
	\end{table}
	
	\section{Experiments}
	
	In this section, we carry on extensive experiments to evaluate the DOAM we proposed. In our work, the main task is to detect occluded prohibited items in X-ray images in security inspection scenario. To the best of our knowledge, no dataset targeting this task has been proposed in the literature, so we only adopt the OPIXray dataset in the experiments. \textbf{First}, we verify that DOAM outperforms all the attention mechanisms mentioned above, over different categories and different occlusion levels. \textbf{Second}, we perform ablation experiments to thoroughly evaluate the effectiveness of DOAM. \textbf{Third}, we demonstrate the general applicability of DOAM across different architectures and the effectiveness after DOAM-integrated. \textbf{Finally}, we apply the Grad-CAM \cite{selvaraju2017grad} to visualize the attention mechanism of DOAM.
	
	\textbf{Evaluation strategy:} All experiments are carried on the OPIXray dataset. In experiments of comparing with different attention mechanisms over different occlusion levels, every model is trained by training set data in Tab. \ref{data_table} and tested on OL1, OL2 and OL3 in Tab. \ref{data_cate_table} respectively. In any other experiments, every model is trained by training set data and tested by the testing set data in Tab. \ref{data_table}.
	
	\textbf{Baseline Detail:} In experiments of comparing with different attention mechanisms, we respectively plug DOAM and each of the attention modules into SSD \cite{liu2016ssd} and report the performances of SSD \cite{liu2016ssd} and these integrated networks. In our experiments, these attention modules are added to the backbone (VGG16) of SSD. More specifically, they are inserted behind the max pooling layer where the feature map is scaled to half. In ablation study, we plug each sub-module of DOAM into SSD \cite{liu2016ssd} separately and report the performances of SSD \cite{liu2016ssd} and these integrated networks. In experiments of comparing with different detection approaches, we plug DOAM into a number of popular detection networks and report the model performance with or without DOAM in every detection network.
	
	\textbf{Parameter setting:} In all experiments following, all models are optimized by the SGD optimizer and the learning rate is set to 0.0001. The batch size is set to 24 and the momentum and weight decay are set to 0.9 and 0.0005 respectively. We evaluate the mean Average Precision (mAP) of the object detection to measure the performance of the model and the IOU threshold is set to 0.5. We further select the best performance model to calculate the AP of each category to observe the performance improvement in different categories. Furthermore, in order to avoid the influence of image data modification on edge image generation, we do not use any data augmentation methods to expand the data or modify the pixel value of the original image, which helps us to better analyze the impact of edge information.  
	
	\subsection{Comparing with Different Attention Mechanisms}
	We compare three variants of attention mechanisms above, including SE \cite{hu2018squeeze}, Non-local \cite{wang2018non} and DA \cite{fu2019dual}. Tab. \ref{table_attention_catagories} and \ref{Occlusion-result} reports the performances of all models.
	
	\textbf{Object Categories:} As we observed from Tab. \ref{table_attention_catagories}, the DOAM-integrated model outperforms SSD \cite{liu2016ssd} by $3.12\%$. Besides, DOAM outperforms SE \cite{hu2018squeeze}, Non-local \cite{wang2018non} and DA \cite{fu2019dual}, by $2.16\%$, $2.60\%$, $2.05\%$, respectively. Moreover, Tab. \ref{table_attention_catagories} shows the improvement of DOAM is mainly reflected in Straight Knife, Folding Knife and Utility Knife, all of which are with the high level occlusion. Especially for Straight Knife, which is the category with highest level occlusion, DOAM outperforms SSD \cite{liu2016ssd} by an impressive amount of $6.48\%$ and Non-local \cite{wang2018non} by $5.12\%$. For Scissor, the lightest occlusion category, the performance of DOAM is only improved by $1.71\%$ compared to SSD \cite{liu2016ssd} and similar to Non-local \cite{wang2018non}. It is obvious that DOAM surpasses these current popular attention mechanisms over different categories. 
	
\begin{figure}[h]
		\centering
		\includegraphics[width=0.9\linewidth]{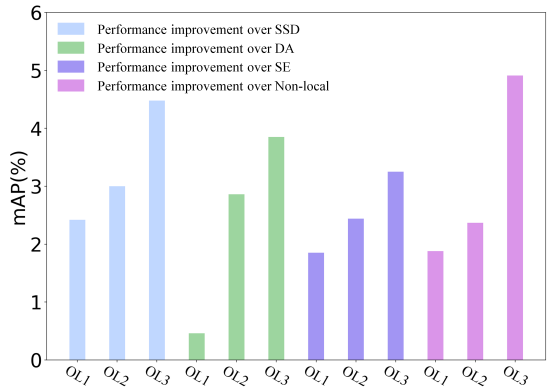}
		\caption{The amount changes of performance improvement of DOAM over different models with occlusion level increasing.}
		\Description{}\label{improve}
	\end{figure}

	\textbf{Object Occlusion Levels:} The experimental results are shown in Tab. \ref{level-table}. Further, Fig. \ref{improve} is drawn according to Tab. \ref{level-table} to illustrate the effectiveness of DOAM to occluded object detection in X-ray images more clearly. In Fig. \ref{improve}, we can clearly obtain a conclusion that DOAM can achieve a higher performance than the baseline and other attention mechanisms with the X-ray images suffer a higher level of occlusion. It verifies the effectiveness of DOAM that it has a significant effect on the performance of detecting occluded prohibited items in X-ray images. (Note that in OL3, the performance of "SSD+Non-local" is lower than "SSD". Due to the attention mechanism of Non-local is to capture spatial information by constructing the relations between regions, we conjecture that this type of relation reduces effect when the noises of image increase.)
	
	\begin{table}[h]
		\setlength{\tabcolsep}{13pt}
		\caption{Performance comparison between DOAM and other different attention mechanisms on object occlusion levels.}
		\label{Occlusion-result}
		\begin{tabular}{lccc}
			\toprule
			Method & OL 1 & OL 2 & OL 3\\
			\midrule
			SSD \cite{liu2016ssd} & 75.45 & 69.54 & 66.30\\
			\hline
			SSD+SE \cite{hu2018squeeze} & 76.02 & 70.11 & 67.53\\
			SSD+Non-local \cite{wang2018non} & 75.99 & 70.17 & 65.87\\
			SSD+DA \cite{fu2019dual} & 77.41 & 69.68 & 66.93\\
			\textbf{SSD+DOAM(ours)} & \textbf{77.87} & \textbf{72.45} & \textbf{70.78}\\
			\bottomrule
		\end{tabular}
		\label{level-table}
	\end{table}
	
	\subsection{Ablation Study}\label{sec:ablation-study}
	Tab. \ref{Ablation-result} shows that EG improved the performance by $0.43\%$ compared with the method of simply concatenating the input image and the corresponding edge image without any other operations of EG. We conjecture that it is mainly because the EG has the ability to focus adaptively on the prohibited items we desire to detect by specifically increasing the weight of edge information through the optimization of a loss function, while simply concatenating operates all the objects in the image equally whether the object is we desire to detect or not for feature fusion. 
	
	Besides, model integrating both EG and MA achieves better performance by $0.37\%$ than integrating EG alone, which verifies the effectiveness of MA. Note that we observe prohibited item size is about 10$\times$10 averagely, so we choose 10$\times$10 as the region scale to perceive for each position of the feature map.
	
	We choose three different scales of the regions ($5\times5$, $10\times10$, $15\times15$ respectively), and draw the gated convolutional neural network \cite{yu2019free} $\mathbb{G}$ into MA, to adaptively select the best feature map which generated by operation of average pooling with appropriate pooling size. The experimental results show that after drawing $\mathbb{G}$, the performance improves by $0.9\%$.
	
	\begin{table}[h]
		\setlength{\tabcolsep}{3pt}
		\caption{Ablation studies of DOAM. "C" represents simply concatenate operation, "DOAM-MA" represents DOAM without Material Awareness module, "DOAM-$\mathbb{G}$" represents DOAM without the Gate Convolutional Neural Network.} 
		\begin{tabular}{lcccccc}
			\toprule
			\multicolumn{1}{l}{\multirow{2}{*}{Method}} & \multicolumn{1}{c}{\multirow{2}{*}{mAP}} & \multicolumn{5}{c}{Category}                                                        \\ \cline{3-7} 
			\multicolumn{1}{l}{} 
			& \multicolumn{1}{c}{}
			& \multicolumn{1}{c}{FO} & \multicolumn{1}{c}{ST} & \multicolumn{1}{c}{SC} & \multicolumn{1}{c}{UT} & \multicolumn{1}{c}{MU} \\ \hline
			SSD \cite{liu2016ssd} & 70.89  & 76.91  & 35.02  & 93.41    & 65.87& 83.27 \\
			\hline
			SSD+C  &72.32 & 79.00  & 36.46  & 94.13  & 68.85    & 83.18 \\
			SSD+(DOAM-MA)  & 72.75 & 80.26  & 35.54 & 94.81  & 67.96  & \textbf{85.19}    \\
			SSD+(DOAM-$\mathbb{G}$) & 73.12 & 79.94 & 38.58 & 93.39    & \textbf{69.40}   & 84.28    \\
			\textbf{SSD+DOAM(ours)}   & \textbf{74.01} & \textbf{81.37}   & \textbf{41.50} & \textbf{95.12}    & 68.21  & 83.83 \\          
			\bottomrule
		\end{tabular}\label{Ablation-result}
	\end{table}

	\subsection{Comparing with Different Detection Approaches}  
	
	To further evaluate the effectiveness of DOAM and verify DOAM can be applied to various detection networks, we conduct experiments on the famous detection approaches, SSD \cite{liu2016ssd}, YOLOv3 \cite{redmon2018yolov3} and FCOS \cite{tian2019fcos}. The results are shown in Tab. \ref{detection_approach}.
	
	\begin{table}[h]
		\setlength{\tabcolsep}{2pt}
		\caption{Performance comparison between DOAM-integrated network and baselines for three famous detection approaches.}   
		\begin{tabular}{lcccccc}
			\toprule
			\multicolumn{1}{l}{\multirow{2}{*}{Method}} & \multicolumn{1}{c}{\multirow{2}{*}{mAP}} & \multicolumn{5}{c}{Category} \\ \cline{3-7} 
			\multicolumn{1}{l}{} 
			& \multicolumn{1}{c}{}
			& \multicolumn{1}{c}{FO} & \multicolumn{1}{c}{ST} & \multicolumn{1}{c}{SC} & \multicolumn{1}{c}{UT} & \multicolumn{1}{c}{MU} \\ \hline
			SSD \cite{liu2016ssd} & 70.89  & 76.91  & 35.02  & 93.41    & 65.87& 83.27 \\
			\textbf{SSD+DOAM(ours)}  & \textbf{74.01}  & \textbf{81.37} & \textbf{41.50}  & \textbf{95.12}  & \textbf{68.21}  & \textbf{83.83} \\
			\hline
			YOLOv3 \cite{redmon2018yolov3}  & 78.21 & \textbf{92.53} & 36.02    & \textbf{97.34}   & 70.81   & 94.37 \\
			\textbf{YOLOv3+DOAM(ours)}   & \textbf{79.25}   & 90.23 & \textbf{41.73}    & 96.96  & \textbf{72.12}  & \textbf{95.23}\\
			\hline
			FCOS \cite{tian2019fcos} & 82.02 & 86.41 & 68.47 & 90.22 & 78.39 & 86.60\\
			\textbf{FCOS+DOAM(ours)} & \textbf{82.41} & \textbf{86.71} & \textbf{68.58} & \textbf{90.23} & \textbf{78.84} & \textbf{87.67}\\
			\bottomrule
		\end{tabular}\label{detection_approach}
	\end{table}       
	
	As we can see from Tab. \ref{detection_approach}, the performance of DOAM-integrated networks are improved by 3.12\%, 1.04\% and 0.39\% compared with SSD \cite{liu2016ssd}, YOLOv3 \cite{redmon2018yolov3} and FCOS \cite{tian2019fcos} respectively, which verify that our module can be inserted as a plug-and-play module into most detection networks and receive a better performance. Note that the performances on Folding Knife and Scissor after DOAM-integrated are slightly reduced. We speculate that the reason is that these images of the two categories in the dataset are occluded not seriously. When the occlusion level increases, the attention mechanism pays more attention to the objects occluded highly like straight knives while less attention to the objects occluded slightly, which results in the slight performance degradation for Folding Knife and Scissor.

	\subsection{Attention Visualization Analysis}
	In this section, we visualize the attention map generated in DOAM to observe the effects of DOAM. The attention distribution can be visualized in Fig. \ref{att_vis}. In rows 1 and 3, we select 10 input X-ray images (each category has two images) and show their corresponding attention visualizations in rows 2 and 4. We observe that DOAM could capture edge and region information accurately. For example, in column 4, a red box is marked on a utility knife of the X-ray image (in row 1), and the boundaries of the utility knife are very clear in the attention visualization (in row 2). Moreover, in the first column, a red box is marked on a folding knife and the corresponding attention map (in row 2) highlights most of the areas where the folding knife lies on. In short, these visualizations further demonstrate the effectiveness of capturing edge and region information for improving feature representation in occluded prohibited items detection.
	
	\begin{figure}[h]
		\centering
		\includegraphics[width=\linewidth]{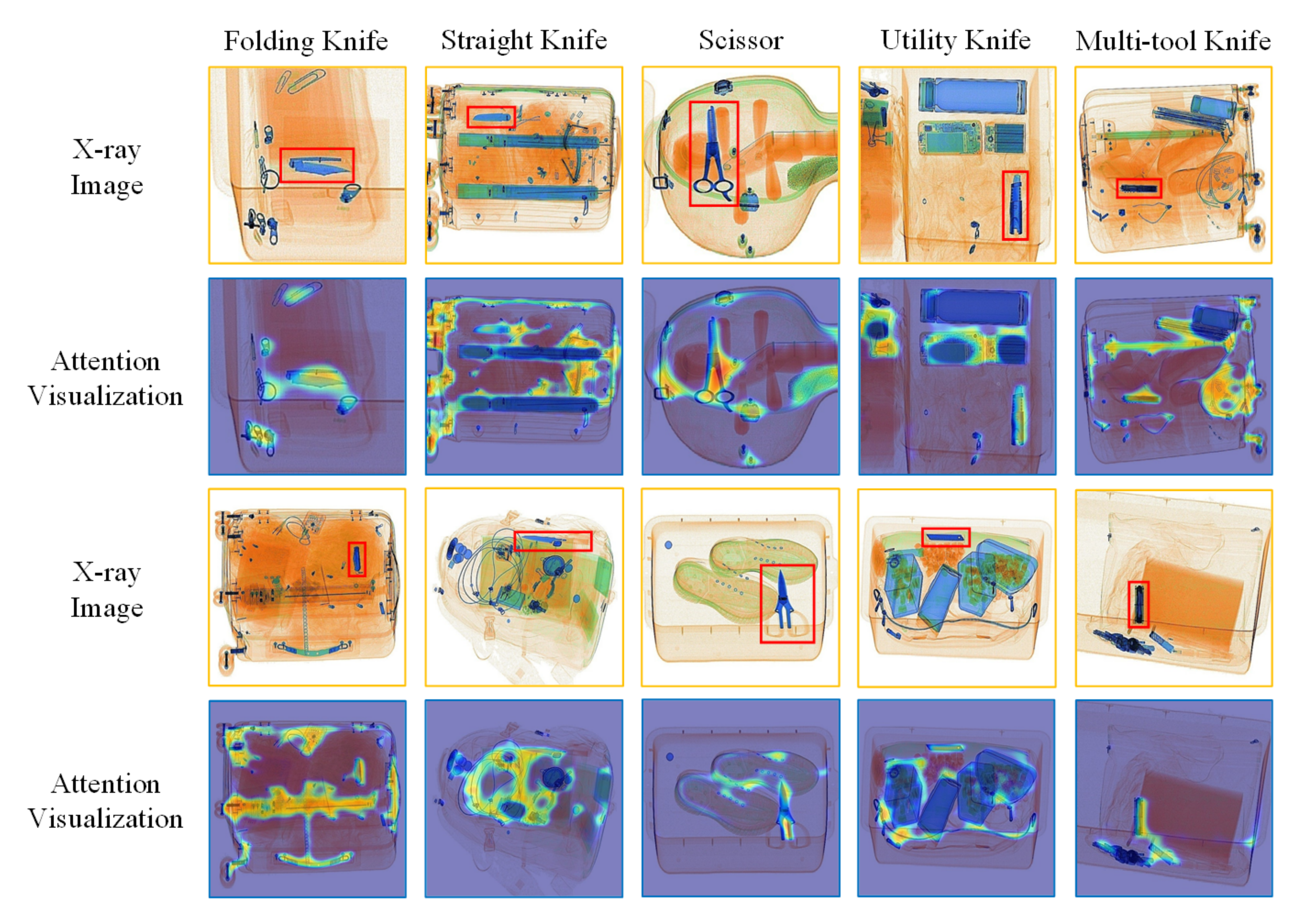}
		\caption{Attention visualization results.}
		\Description{}\label{att_vis}
		\label{att_vis}
	\end{figure}

	\section{Conclusion}
	In this paper, we investigate occluded prohibited items detection in X-ray scanned images, which is a promising application in industry yet remains fewer studied in computer vision. To facilitate research in this field, we contribute the first high-quality object detection dataset for security inspection, named Occluded Prohibited Items X-ray (OPIXray) image benchmark. OPIXray focused on the widely-occurred prohibited item "cutter", annotated manually by professional inspectors from the international airport.
	To deal with the occlusion in X-ray images detection, we propose the De-occlusion Attention Module (DOAM), a plug-and-play module that can be easily inserted into and thus promote most popular detectors. We comprehensively evaluate our module on the OPIXray dataset, and demonstrate that our module can consistently improve the performance of the state-of-the-art detection methods such as SSD, FCOS, etc, and significantly outperforms several widely-used attention mechanisms. In particular, the advantages of DOAM are more significant in the scenarios with higher levels of occlusion, which demonstrates its potential application in real-world inspections.
	
	\section*{Acknowledgment}
	This work was supported by National Natural Science Foundation of China (61872021), Beijing Nova Program of Science and Technology (Z191100001119050), State Key Lab of Software Development Environment (SKLSDE-2020ZX-06) and Fundamental Research Funds for Central Universities (YWF-20-BJ-J-646).
	
	\clearpage
	\bibliographystyle{ACM-Reference-Format}
	\bibliography{OPIXray}


\begin{thebibliography}{30}


\ifx \showCODEN    \undefined \def \showCODEN     #1{\unskip}     \fi
\ifx \showDOI      \undefined \def \showDOI       #1{#1}\fi
\ifx \showISBNx    \undefined \def \showISBNx     #1{\unskip}     \fi
\ifx \showISBNxiii \undefined \def \showISBNxiii  #1{\unskip}     \fi
\ifx \showISSN     \undefined \def \showISSN      #1{\unskip}     \fi
\ifx \showLCCN     \undefined \def \showLCCN      #1{\unskip}     \fi
\ifx \shownote     \undefined \def \shownote      #1{#1}          \fi
\ifx \showarticletitle \undefined \def \showarticletitle #1{#1}   \fi
\ifx \showURL      \undefined \def \showURL       {\relax}        \fi
\providecommand\bibfield[2]{#2}
\providecommand\bibinfo[2]{#2}
\providecommand\natexlab[1]{#1}
\providecommand\showeprint[2][]{arXiv:#2}

\bibitem[\protect\citeauthoryear{Chaudhary, Hazra, and Chaudhary}{Chaudhary
  et~al\mbox{.}}{2019}]%
        {chaudhary2019diagnosis}
\bibfield{author}{\bibinfo{person}{Arjun Chaudhary}, \bibinfo{person}{Abhishek
  Hazra}, {and} \bibinfo{person}{Prakash Chaudhary}.}
  \bibinfo{year}{2019}\natexlab{}.
\newblock \showarticletitle{Diagnosis of Chest Diseases in X-Ray images using
  Deep Convolutional Neural Network}. In \bibinfo{booktitle}{\emph{2019 10th
  International Conference on Computing, Communication and Networking
  Technologies (ICCCNT)}}. IEEE, \bibinfo{pages}{1--6}.
\newblock


\bibitem[\protect\citeauthoryear{Chen, Zhang, Xiao, Nie, Shao, Liu, and
  Chua}{Chen et~al\mbox{.}}{2017}]%
        {chen2017sca}
\bibfield{author}{\bibinfo{person}{Long Chen}, \bibinfo{person}{Hanwang Zhang},
  \bibinfo{person}{Jun Xiao}, \bibinfo{person}{Liqiang Nie},
  \bibinfo{person}{Jian Shao}, \bibinfo{person}{Wei Liu}, {and}
  \bibinfo{person}{Tat-Seng Chua}.} \bibinfo{year}{2017}\natexlab{}.
\newblock \showarticletitle{Sca-cnn: Spatial and channel-wise attention in
  convolutional networks for image captioning}. In
  \bibinfo{booktitle}{\emph{Proceedings of the IEEE conference on computer
  vision and pattern recognition}}. \bibinfo{pages}{5659--5667}.
\newblock


\bibitem[\protect\citeauthoryear{Cheng, Li, Dai, Wu, He, and Hauptmann}{Cheng
  et~al\mbox{.}}{2019}]%
        {cheng2019improving}
\bibfield{author}{\bibinfo{person}{Zhi-Qi Cheng}, \bibinfo{person}{Jun-Xiu Li},
  \bibinfo{person}{Qi Dai}, \bibinfo{person}{Xiao Wu}, \bibinfo{person}{Jun-Yan
  He}, {and} \bibinfo{person}{Alexander~G Hauptmann}.}
  \bibinfo{year}{2019}\natexlab{}.
\newblock \showarticletitle{Improving the learning of multi-column
  convolutional neural network for crowd counting}. In
  \bibinfo{booktitle}{\emph{Proceedings of the 27th ACM International
  Conference on Multimedia}}. \bibinfo{pages}{1897--1906}.
\newblock


\bibitem[\protect\citeauthoryear{Fu, Liu, Tian, Li, Bao, Fang, and Lu}{Fu
  et~al\mbox{.}}{2019}]%
        {fu2019dual}
\bibfield{author}{\bibinfo{person}{Jun Fu}, \bibinfo{person}{Jing Liu},
  \bibinfo{person}{Haijie Tian}, \bibinfo{person}{Yong Li},
  \bibinfo{person}{Yongjun Bao}, \bibinfo{person}{Zhiwei Fang}, {and}
  \bibinfo{person}{Hanqing Lu}.} \bibinfo{year}{2019}\natexlab{}.
\newblock \showarticletitle{Dual attention network for scene segmentation}. In
  \bibinfo{booktitle}{\emph{Proceedings of the IEEE Conference on Computer
  Vision and Pattern Recognition}}. \bibinfo{pages}{3146--3154}.
\newblock


\bibitem[\protect\citeauthoryear{Ge, Li, Ye, and Luo}{Ge et~al\mbox{.}}{2017}]%
        {ge2017detecting}
\bibfield{author}{\bibinfo{person}{Shiming Ge}, \bibinfo{person}{Jia Li},
  \bibinfo{person}{Qiting Ye}, {and} \bibinfo{person}{Zhao Luo}.}
  \bibinfo{year}{2017}\natexlab{}.
\newblock \showarticletitle{Detecting masked faces in the wild with lle-cnns}.
  In \bibinfo{booktitle}{\emph{Proceedings of the IEEE Conference on Computer
  Vision and Pattern Recognition}}. \bibinfo{pages}{2682--2690}.
\newblock


\bibitem[\protect\citeauthoryear{Guo, Tang, Zhu, Fan, Ai, Song, Liang, and
  Yang}{Guo et~al\mbox{.}}{2019}]%
        {guo2019improved}
\bibfield{author}{\bibinfo{person}{Shuai Guo}, \bibinfo{person}{Songyuan Tang},
  \bibinfo{person}{Jianjun Zhu}, \bibinfo{person}{Jingfan Fan},
  \bibinfo{person}{Danni Ai}, \bibinfo{person}{Hong Song},
  \bibinfo{person}{Ping Liang}, {and} \bibinfo{person}{Jian Yang}.}
  \bibinfo{year}{2019}\natexlab{}.
\newblock \showarticletitle{Improved U-Net for Guidewire Tip Segmentation in
  X-ray Fluoroscopy Images}. In \bibinfo{booktitle}{\emph{Proceedings of the
  2019 3rd International Conference on Advances in Image Processing}}.
  \bibinfo{pages}{55--59}.
\newblock


\bibitem[\protect\citeauthoryear{Hu, Shen, and Sun}{Hu et~al\mbox{.}}{2018}]%
        {hu2018squeeze}
\bibfield{author}{\bibinfo{person}{Jie Hu}, \bibinfo{person}{Li Shen}, {and}
  \bibinfo{person}{Gang Sun}.} \bibinfo{year}{2018}\natexlab{}.
\newblock \showarticletitle{Squeeze-and-excitation networks}. In
  \bibinfo{booktitle}{\emph{Proceedings of the IEEE conference on computer
  vision and pattern recognition}}. \bibinfo{pages}{7132--7141}.
\newblock


\bibitem[\protect\citeauthoryear{Huang, Wang, Chen, Xu, Tang, and Mu}{Huang
  et~al\mbox{.}}{2019}]%
        {huang2019modeling}
\bibfield{author}{\bibinfo{person}{Shengling Huang}, \bibinfo{person}{Xin
  Wang}, \bibinfo{person}{Yifan Chen}, \bibinfo{person}{Jie Xu},
  \bibinfo{person}{Tian Tang}, {and} \bibinfo{person}{Baozhong Mu}.}
  \bibinfo{year}{2019}\natexlab{}.
\newblock \showarticletitle{Modeling and quantitative analysis of X-ray
  transmission and backscatter imaging aimed at security inspection}.
\newblock \bibinfo{journal}{\emph{Optics express}} \bibinfo{volume}{27},
  \bibinfo{number}{2} (\bibinfo{year}{2019}), \bibinfo{pages}{337--349}.
\newblock


\bibitem[\protect\citeauthoryear{LeCun, Bengio, and Hinton}{LeCun
  et~al\mbox{.}}{2015}]%
        {lecun2015deep}
\bibfield{author}{\bibinfo{person}{Yann LeCun}, \bibinfo{person}{Yoshua
  Bengio}, {and} \bibinfo{person}{Geoffrey Hinton}.}
  \bibinfo{year}{2015}\natexlab{}.
\newblock \showarticletitle{Deep learning}.
\newblock \bibinfo{journal}{\emph{nature}} \bibinfo{volume}{521},
  \bibinfo{number}{7553} (\bibinfo{year}{2015}), \bibinfo{pages}{436--444}.
\newblock


\bibitem[\protect\citeauthoryear{Liu, Anguelov, Erhan, Szegedy, Reed, Fu, and
  Berg}{Liu et~al\mbox{.}}{2016}]%
        {liu2016ssd}
\bibfield{author}{\bibinfo{person}{Wei Liu}, \bibinfo{person}{Dragomir
  Anguelov}, \bibinfo{person}{Dumitru Erhan}, \bibinfo{person}{Christian
  Szegedy}, \bibinfo{person}{Scott Reed}, \bibinfo{person}{Cheng-Yang Fu},
  {and} \bibinfo{person}{Alexander~C Berg}.} \bibinfo{year}{2016}\natexlab{}.
\newblock \showarticletitle{Ssd: Single shot multibox detector}. In
  \bibinfo{booktitle}{\emph{European conference on computer vision}}. Springer,
  \bibinfo{pages}{21--37}.
\newblock


\bibitem[\protect\citeauthoryear{Lu and Tong}{Lu and Tong}{2019}]%
        {lu2019towards}
\bibfield{author}{\bibinfo{person}{Jianjie Lu} {and} \bibinfo{person}{Kai-yu
  Tong}.} \bibinfo{year}{2019}\natexlab{}.
\newblock \showarticletitle{Towards to Reasonable Decision Basis in Automatic
  Bone X-Ray Image Classification: A Weakly-Supervised Approach}. In
  \bibinfo{booktitle}{\emph{Proceedings of the AAAI Conference on Artificial
  Intelligence}}, Vol.~\bibinfo{volume}{33}. \bibinfo{pages}{9985--9986}.
\newblock


\bibitem[\protect\citeauthoryear{Mery, Riffo, Zscherpel, Mondrag{\'o}n, Lillo,
  Zuccar, Lobel, and Carrasco}{Mery et~al\mbox{.}}{2015}]%
        {mery2015gdxray}
\bibfield{author}{\bibinfo{person}{Domingo Mery}, \bibinfo{person}{Vladimir
  Riffo}, \bibinfo{person}{Uwe Zscherpel}, \bibinfo{person}{German
  Mondrag{\'o}n}, \bibinfo{person}{Iv{\'a}n Lillo}, \bibinfo{person}{Irene
  Zuccar}, \bibinfo{person}{Hans Lobel}, {and} \bibinfo{person}{Miguel
  Carrasco}.} \bibinfo{year}{2015}\natexlab{}.
\newblock \showarticletitle{GDXray: The database of X-ray images for
  nondestructive testing}.
\newblock \bibinfo{journal}{\emph{Journal of Nondestructive Evaluation}}
  \bibinfo{volume}{34}, \bibinfo{number}{4} (\bibinfo{year}{2015}),
  \bibinfo{pages}{42}.
\newblock


\bibitem[\protect\citeauthoryear{Miao, Xie, Wan, Su, Liu, Jiao, and Ye}{Miao
  et~al\mbox{.}}{2019}]%
        {miao2019sixray}
\bibfield{author}{\bibinfo{person}{Caijing Miao}, \bibinfo{person}{Lingxi Xie},
  \bibinfo{person}{Fang Wan}, \bibinfo{person}{Chi Su}, \bibinfo{person}{Hongye
  Liu}, \bibinfo{person}{Jianbin Jiao}, {and} \bibinfo{person}{Qixiang Ye}.}
  \bibinfo{year}{2019}\natexlab{}.
\newblock \showarticletitle{Sixray: A large-scale security inspection x-ray
  benchmark for prohibited item discovery in overlapping images}. In
  \bibinfo{booktitle}{\emph{Proceedings of the IEEE Conference on Computer
  Vision and Pattern Recognition}}. \bibinfo{pages}{2119--2128}.
\newblock


\bibitem[\protect\citeauthoryear{Peng, Yang, Wang, Wu, and Huang}{Peng
  et~al\mbox{.}}{2019}]%
        {peng2019cra}
\bibfield{author}{\bibinfo{person}{Liang Peng}, \bibinfo{person}{Yang Yang},
  \bibinfo{person}{Zheng Wang}, \bibinfo{person}{Xiao Wu}, {and}
  \bibinfo{person}{Zi Huang}.} \bibinfo{year}{2019}\natexlab{}.
\newblock \showarticletitle{CRA-Net: Composed Relation Attention Network for
  Visual Question Answering}. In \bibinfo{booktitle}{\emph{Proceedings of the
  27th ACM International Conference on Multimedia}}.
  \bibinfo{pages}{1202--1210}.
\newblock


\bibitem[\protect\citeauthoryear{Redmon and Farhadi}{Redmon and
  Farhadi}{2018}]%
        {redmon2018yolov3}
\bibfield{author}{\bibinfo{person}{Joseph Redmon} {and} \bibinfo{person}{Ali
  Farhadi}.} \bibinfo{year}{2018}\natexlab{}.
\newblock \showarticletitle{Yolov3: An incremental improvement}.
\newblock \bibinfo{journal}{\emph{arXiv preprint arXiv:1804.02767}}
  (\bibinfo{year}{2018}).
\newblock


\bibitem[\protect\citeauthoryear{Selvaraju, Cogswell, Das, Vedantam, Parikh,
  and Batra}{Selvaraju et~al\mbox{.}}{2017}]%
        {selvaraju2017grad}
\bibfield{author}{\bibinfo{person}{Ramprasaath~R Selvaraju},
  \bibinfo{person}{Michael Cogswell}, \bibinfo{person}{Abhishek Das},
  \bibinfo{person}{Ramakrishna Vedantam}, \bibinfo{person}{Devi Parikh}, {and}
  \bibinfo{person}{Dhruv Batra}.} \bibinfo{year}{2017}\natexlab{}.
\newblock \showarticletitle{Grad-cam: Visual explanations from deep networks
  via gradient-based localization}. In \bibinfo{booktitle}{\emph{Proceedings of
  the IEEE international conference on computer vision}}.
  \bibinfo{pages}{618--626}.
\newblock


\bibitem[\protect\citeauthoryear{Song, Gong, Li, Liu, and Liu}{Song
  et~al\mbox{.}}{2019}]%
        {song2019occlusion}
\bibfield{author}{\bibinfo{person}{Lingxue Song}, \bibinfo{person}{Dihong
  Gong}, \bibinfo{person}{Zhifeng Li}, \bibinfo{person}{Changsong Liu}, {and}
  \bibinfo{person}{Wei Liu}.} \bibinfo{year}{2019}\natexlab{}.
\newblock \showarticletitle{Occlusion Robust Face Recognition Based on Mask
  Learning With Pairwise Differential Siamese Network}. In
  \bibinfo{booktitle}{\emph{Proceedings of the IEEE International Conference on
  Computer Vision}}. \bibinfo{pages}{773--782}.
\newblock


\bibitem[\protect\citeauthoryear{Sun, Jin, and Li}{Sun et~al\mbox{.}}{2019}]%
        {sun2019attention}
\bibfield{author}{\bibinfo{person}{Xie Sun}, \bibinfo{person}{Lu Jin}, {and}
  \bibinfo{person}{Zechao Li}.} \bibinfo{year}{2019}\natexlab{}.
\newblock \showarticletitle{Attention-Aware Feature Pyramid Ordinal Hashing for
  Image Retrieval}.
\newblock In \bibinfo{booktitle}{\emph{Proceedings of the ACM Multimedia Asia
  on ZZZ}}. \bibinfo{pages}{1--6}.
\newblock


\bibitem[\protect\citeauthoryear{Tang, Jin, Li, and Gao}{Tang
  et~al\mbox{.}}{2015}]%
        {tang2015rgb}
\bibfield{author}{\bibinfo{person}{Jinhui Tang}, \bibinfo{person}{Lu Jin},
  \bibinfo{person}{Zechao Li}, {and} \bibinfo{person}{Shenghua Gao}.}
  \bibinfo{year}{2015}\natexlab{}.
\newblock \showarticletitle{RGB-D object recognition via incorporating latent
  data structure and prior knowledge}.
\newblock \bibinfo{journal}{\emph{IEEE Transactions on Multimedia}}
  \bibinfo{volume}{17}, \bibinfo{number}{11} (\bibinfo{year}{2015}),
  \bibinfo{pages}{1899--1908}.
\newblock


\bibitem[\protect\citeauthoryear{Tian, Shen, Chen, and He}{Tian
  et~al\mbox{.}}{2019}]%
        {tian2019fcos}
\bibfield{author}{\bibinfo{person}{Zhi Tian}, \bibinfo{person}{Chunhua Shen},
  \bibinfo{person}{Hao Chen}, {and} \bibinfo{person}{Tong He}.}
  \bibinfo{year}{2019}\natexlab{}.
\newblock \showarticletitle{Fcos: Fully convolutional one-stage object
  detection}. In \bibinfo{booktitle}{\emph{Proceedings of the IEEE
  International Conference on Computer Vision}}. \bibinfo{pages}{9627--9636}.
\newblock


\bibitem[\protect\citeauthoryear{Vaswani, Shazeer, Parmar, Uszkoreit, Jones,
  Gomez, Kaiser, and Polosukhin}{Vaswani et~al\mbox{.}}{2017}]%
        {vaswani2017attention}
\bibfield{author}{\bibinfo{person}{Ashish Vaswani}, \bibinfo{person}{Noam
  Shazeer}, \bibinfo{person}{Niki Parmar}, \bibinfo{person}{Jakob Uszkoreit},
  \bibinfo{person}{Llion Jones}, \bibinfo{person}{Aidan~N Gomez},
  \bibinfo{person}{{\L}ukasz Kaiser}, {and} \bibinfo{person}{Illia
  Polosukhin}.} \bibinfo{year}{2017}\natexlab{}.
\newblock \showarticletitle{Attention is all you need}. In
  \bibinfo{booktitle}{\emph{Advances in neural information processing
  systems}}. \bibinfo{pages}{5998--6008}.
\newblock


\bibitem[\protect\citeauthoryear{Wang, Li, Hou, and Li}{Wang
  et~al\mbox{.}}{2016}]%
        {wang2016action}
\bibfield{author}{\bibinfo{person}{Pichao Wang}, \bibinfo{person}{Zhaoyang Li},
  \bibinfo{person}{Yonghong Hou}, {and} \bibinfo{person}{Wanqing Li}.}
  \bibinfo{year}{2016}\natexlab{}.
\newblock \showarticletitle{Action recognition based on joint trajectory maps
  using convolutional neural networks}. In
  \bibinfo{booktitle}{\emph{Proceedings of the 24th ACM international
  conference on Multimedia}}. \bibinfo{pages}{102--106}.
\newblock


\bibitem[\protect\citeauthoryear{Wang, Girshick, Gupta, and He}{Wang
  et~al\mbox{.}}{2018a}]%
        {wang2018non}
\bibfield{author}{\bibinfo{person}{Xiaolong Wang}, \bibinfo{person}{Ross
  Girshick}, \bibinfo{person}{Abhinav Gupta}, {and} \bibinfo{person}{Kaiming
  He}.} \bibinfo{year}{2018}\natexlab{a}.
\newblock \showarticletitle{Non-local neural networks}. In
  \bibinfo{booktitle}{\emph{Proceedings of the IEEE conference on computer
  vision and pattern recognition}}. \bibinfo{pages}{7794--7803}.
\newblock


\bibitem[\protect\citeauthoryear{Wang, Xiao, Jiang, Shao, Sun, and Shen}{Wang
  et~al\mbox{.}}{2018b}]%
        {wang2018repulsion}
\bibfield{author}{\bibinfo{person}{Xinlong Wang}, \bibinfo{person}{Tete Xiao},
  \bibinfo{person}{Yuning Jiang}, \bibinfo{person}{Shuai Shao},
  \bibinfo{person}{Jian Sun}, {and} \bibinfo{person}{Chunhua Shen}.}
  \bibinfo{year}{2018}\natexlab{b}.
\newblock \showarticletitle{Repulsion loss: Detecting pedestrians in a crowd}.
  In \bibinfo{booktitle}{\emph{Proceedings of the IEEE Conference on Computer
  Vision and Pattern Recognition}}. \bibinfo{pages}{7774--7783}.
\newblock


\bibitem[\protect\citeauthoryear{Yang, Luo, Qian, Tai, Zhang, and Xu}{Yang
  et~al\mbox{.}}{2016}]%
        {yang2016nuclear}
\bibfield{author}{\bibinfo{person}{Jian Yang}, \bibinfo{person}{Lei Luo},
  \bibinfo{person}{Jianjun Qian}, \bibinfo{person}{Ying Tai},
  \bibinfo{person}{Fanlong Zhang}, {and} \bibinfo{person}{Yong Xu}.}
  \bibinfo{year}{2016}\natexlab{}.
\newblock \showarticletitle{Nuclear norm based matrix regression with
  applications to face recognition with occlusion and illumination changes}.
\newblock \bibinfo{journal}{\emph{IEEE transactions on pattern analysis and
  machine intelligence}} \bibinfo{volume}{39}, \bibinfo{number}{1}
  (\bibinfo{year}{2016}), \bibinfo{pages}{156--171}.
\newblock


\bibitem[\protect\citeauthoryear{Yao, She, Zhao, Liang, Lai, and Yang}{Yao
  et~al\mbox{.}}{2019}]%
        {yao2019attention}
\bibfield{author}{\bibinfo{person}{Xingxu Yao}, \bibinfo{person}{Dongyu She},
  \bibinfo{person}{Sicheng Zhao}, \bibinfo{person}{Jie Liang},
  \bibinfo{person}{Yu-Kun Lai}, {and} \bibinfo{person}{Jufeng Yang}.}
  \bibinfo{year}{2019}\natexlab{}.
\newblock \showarticletitle{Attention-aware polarity sensitive embedding for
  affective image retrieval}. In \bibinfo{booktitle}{\emph{Proceedings of the
  IEEE International Conference on Computer Vision}}.
  \bibinfo{pages}{1140--1150}.
\newblock


\bibitem[\protect\citeauthoryear{Yu, Lin, Yang, Shen, Lu, and Huang}{Yu
  et~al\mbox{.}}{2019}]%
        {yu2019free}
\bibfield{author}{\bibinfo{person}{Jiahui Yu}, \bibinfo{person}{Zhe Lin},
  \bibinfo{person}{Jimei Yang}, \bibinfo{person}{Xiaohui Shen},
  \bibinfo{person}{Xin Lu}, {and} \bibinfo{person}{Thomas~S Huang}.}
  \bibinfo{year}{2019}\natexlab{}.
\newblock \showarticletitle{Free-form image inpainting with gated convolution}.
  In \bibinfo{booktitle}{\emph{Proceedings of the IEEE International Conference
  on Computer Vision}}. \bibinfo{pages}{4471--4480}.
\newblock


\bibitem[\protect\citeauthoryear{Zha, Liu, Yang, and Zhang}{Zha
  et~al\mbox{.}}{2019}]%
        {zha2019spatiotemporal}
\bibfield{author}{\bibinfo{person}{Zheng-Jun Zha}, \bibinfo{person}{Jiawei
  Liu}, \bibinfo{person}{Tianhao Yang}, {and} \bibinfo{person}{Yongdong
  Zhang}.} \bibinfo{year}{2019}\natexlab{}.
\newblock \showarticletitle{Spatiotemporal-Textual Co-Attention Network for
  Video Question Answering}.
\newblock \bibinfo{journal}{\emph{ACM Transactions on Multimedia Computing,
  Communications, and Applications (TOMM)}} \bibinfo{volume}{15},
  \bibinfo{number}{2s} (\bibinfo{year}{2019}), \bibinfo{pages}{1--18}.
\newblock


\bibitem[\protect\citeauthoryear{Zhang, Wen, Bian, Lei, and Li}{Zhang
  et~al\mbox{.}}{2018}]%
        {zhang2018occlusion}
\bibfield{author}{\bibinfo{person}{Shifeng Zhang}, \bibinfo{person}{Longyin
  Wen}, \bibinfo{person}{Xiao Bian}, \bibinfo{person}{Zhen Lei}, {and}
  \bibinfo{person}{Stan~Z Li}.} \bibinfo{year}{2018}\natexlab{}.
\newblock \showarticletitle{Occlusion-aware R-CNN: detecting pedestrians in a
  crowd}. In \bibinfo{booktitle}{\emph{Proceedings of the European Conference
  on Computer Vision (ECCV)}}. \bibinfo{pages}{637--653}.
\newblock


\bibitem[\protect\citeauthoryear{Zhou and Yuan}{Zhou and Yuan}{2018}]%
        {zhou2018bi}
\bibfield{author}{\bibinfo{person}{Chunluan Zhou} {and}
  \bibinfo{person}{Junsong Yuan}.} \bibinfo{year}{2018}\natexlab{}.
\newblock \showarticletitle{Bi-box regression for pedestrian detection and
  occlusion estimation}. In \bibinfo{booktitle}{\emph{Proceedings of the
  European Conference on Computer Vision (ECCV)}}. \bibinfo{pages}{135--151}.
\newblock


\end{thebibliography}
	
\end{document}